\documentclass[letterpaper, conference]{IEEEtran}
\IEEEoverridecommandlockouts
\usepackage{cite}
\usepackage{amsmath,amssymb,amsfonts}
\usepackage{algorithmic}
\usepackage{graphicx}
\usepackage{textcomp}
\usepackage{xcolor}
\usepackage{fancyhdr}
\usepackage{colortbl}
\usepackage[colorlinks,urlcolor=blue,linkcolor=blue,citecolor=blue]{hyperref}
\usepackage{lipsum}

\def\BibTeX{{\rm B\kern-.05em{\sc i\kern-.025em b}\kern-.08em
    T\kern-.1667em\lower.7ex\hbox{E}\kern-.125emX}}
    
\usepackage{tgpagella}
\usepackage{todonotes}
\usepackage{balance}
\usepackage{tikz}
\usetikzlibrary{arrows.meta, positioning, fit}
\usepackage{colortbl,xcolor}
\usepackage{subcaption}
\usepackage{comment}
\usepackage{amsmath,amssymb}
\usepackage{graphicx}
\usepackage{xcolor}
\usepackage{algorithm}
\usepackage{algorithmic}
\usepackage{balance}
\usepackage{todonotes}
\usepackage{comment}
\usepackage{tgpagella}
\usepackage{tikz}
\usetikzlibrary{arrows.meta, positioning, fit}
\usepackage{colortbl,xcolor}
\usepackage{subcaption}

\begin{document}

\title{\textbf{AI-CARE}: \textbf{C}arbon-\textbf{A}ware \textbf{R}eporting \textbf{E}valuation Metric for \textbf{AI} Models}

\author{
\IEEEauthorblockN{KC Santosh,
Srikanth Baride,
Rodrigue Rizk}

\IEEEauthorblockA{
USD Artificial Intelligence Research, Department of Computer Science\\
University of South Dakota\\
Vermillion, SD 57069, USA\\
Email: \{kc.santosh, srikanth.baride, rodrigue.rizk\}@usd.edu}
}

\maketitle

\thispagestyle{empty}
\pagestyle{empty}

\begin{abstract}

As machine learning (ML) continues its rapid expansion, the environmental cost of model training and inference has become a critical societal concern. Existing benchmarks overwhelmingly focus on standard performance metrics such as accuracy, BLEU, or mAP, while largely ignoring energy consumption and carbon emissions. This single-objective evaluation paradigm is increasingly misaligned with the practical requirements of large-scale deployment, particularly in energy-constrained environments such as mobile devices, developing regions, and climate-aware enterprises. In this paper, we propose AI-CARE, an evaluation tool for reporting energy consumption, and carbon emissions of ML models. In addition, we introduce the carbon-performance tradeoff curve, an interpretable tool that visualizes the Pareto frontier between performance and carbon cost. We demonstrate, through theoretical analysis and empirical validation on representative ML workloads, that carbon-aware benchmarking changes the relative ranking of models and encourages architectures that are simultaneously accurate and environmentally responsible. Our proposal aims to shift the research community toward transparent, multi-objective evaluation and align ML progress with global sustainability goals. The tool and documentation are available at \url{https://github.com/USD-AI-ResearchLab/ai-care}.

\end{abstract}

\section{Introduction}


Machine learning (ML) evaluation has traditionally focused on task-level
performance metrics such as accuracy, F1-score, or mean average precision~\cite{Schwartz2020GreenAI}. While these metrics are essential for assessing
predictive quality, they provide no visibility into the energy consumption or
carbon emissions incurred during model inference and deployment. This limitation
has become increasingly pronounced with the widespread adoption of large-scale
models, including large language models (LLMs) and agentic AI systems, whose
training and deployment are associated with substantial computational and
energy demands that raise environmental concerns\footnote{TEDx Talk: \href{https://www.youtube.com/watch?v=J9dZV2EAuUU}{https://www.youtube.com/watch?v=J9dZV2EAuUU} --- Building sustainable AI for all.} \cite{santosh2025carbonneutralhumanairethinking}.

Despite growing awareness of sustainable and energy-efficient AI, energy and
carbon information is rarely reported in a standardized or comparable manner \cite{santosh2025carbonneutralhumanairethinking}.
Consequently, models with similar predictive performance but vastly different
deployment footprints are often treated as equivalent. This paper focuses on
enabling transparent and reproducible reporting of such costs through a
dedicated evaluation tool.

To address this gap, we introduce AI-CARE (\textbf{C}arbon-\textbf{A}ware \textbf{R}eporting \textbf{E}valuation Tool for \textbf{AI}), a general evaluation tool designed to standardize the
reporting of energy consumption and carbon emissions alongside task
performance. AI-CARE is intended to complement existing accuracy-centric
benchmarks by making deployment-related costs explicit, measurable, and
comparable across models and experimental settings. The framework is agnostic
to model architecture, learning algorithm, and training procedure, and can be
applied across a wide range of ML workflows. The main contributions of this work are summarized as follows:

\begin{itemize}
    \item We present a \emph{general carbon-aware evaluation framework} for
    measuring and reporting energy consumption and derived carbon emissions of
    ML models under controlled experimental conditions.

    \item We introduce the \emph{carbon-performance tradeoff curve} as a
    standardized visualization that exposes empirical trade-offs between
    predictive performance and carbon cost, enabling transparent comparison
    across models and configurations.

    \item We define a \emph{scalar carbon-aware score} to support scenarios in
    which a single ranking is required, while preserving the underlying
    multi-objective nature of accuracy–performance trade-offs.

    \item We release AI-CARE as \emph{open-source software} to encourage
    reproducible, deployment-aware benchmarking and to facilitate integration
    with existing evaluation pipelines.
\end{itemize}

\section{Energy Is All You Forgot}

Standard performance evaluation metrics (e.g., accuracy) obscure a critical
dimension of real-world model deployment: operational efficiency. In practice,
ML models are executed repeatedly, often at scale, where even
small differences in per-inference energy consumption can accumulate into
substantial economic and environmental cost. Conventional benchmarks \cite{Deng2009ImageNet,Wang2018GLUE,Schwartz2020GreenAI} emphasize
predictive performance while largely omitting energy, latency, or carbon
considerations, making deployment-relevant trade-offs effectively invisible.

As a result, accuracy-only rankings systematically favor larger and more
carbon-intensive models, even when their performance gains over lightweight
alternatives are marginal \cite{Strubell2019EnergyAP,Schwartz2020GreenAI}.
Without explicit reporting of energy and carbon costs, practitioners lack the
information needed to make informed, sustainability-aware model selection
decisions.

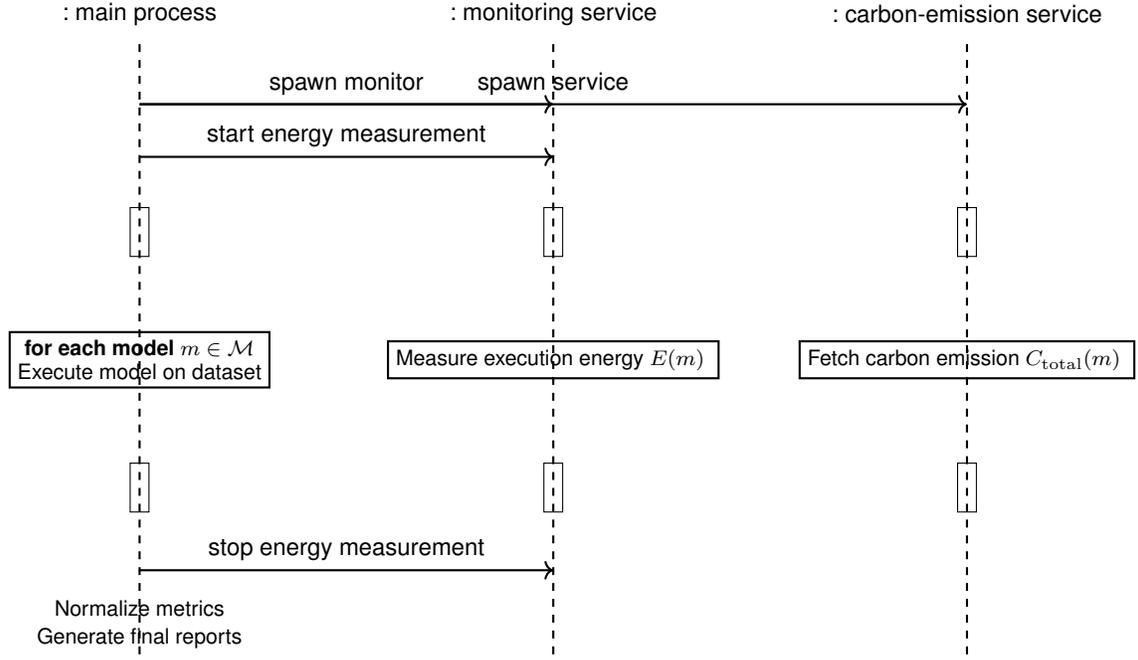
\begin{figure*}[!t]
\centering
\begin{tikzpicture}[
    every node/.style={font=\small\sffamily},
    every label/.style={font=\small\sffamily},
    lifeline/.style={dashed, thick},
    header/.style={align=center},
    activation/.style={draw, rectangle, minimum width=0.25cm, minimum height=0.65cm},
    msg/.style={->, thick},
    loopbox/.style={draw, thick, inner sep=3pt, align=center, font=\footnotesize\sffamily}
]

\node[header] at (0,0)   {: main process};
\node[header] at (5.5,0) {: monitoring service};
\node[header] at (11,0)  {: carbon-emission service};

\draw[lifeline] (0,-0.4) -- (0,-8.6);
\draw[lifeline] (5.5,-0.4) -- (5.5,-8.6);
\draw[lifeline] (11,-0.4) -- (11,-8.6);

\node[activation] at (0,-2.9)  {};
\node[activation] at (5.5,-2.9){};
\node[activation] at (11,-2.9) {};

\draw[msg] (0,-1.2) -- node[above]{spawn monitor} (5.5,-1.2);
\draw[msg] (0,-1.2) -- node[above]{spawn service} (11,-1.2);

\draw[msg] (0,-1.9) -- node[above]{start energy measurement} (5.5,-1.9);

\node[loopbox] at (0,-4.6) {%
\textbf{for each model $m \in \mathcal{M}$}\\
Execute model on dataset
};

\node[loopbox] at (5.5,-4.6) {%
Measure execution energy $E(m)$
};

\node[loopbox] at (11,-4.6) {%
Fetch carbon emission $C_{\mathrm{total}}(m)$
};

\node[activation] at (0,-6.3)  {};
\node[activation] at (5.5,-6.3){};
\node[activation] at (11,-6.3) {};

\draw[msg] (0,-7.4) -- node[above]{stop energy measurement} (5.5,-7.4);

\node[align=left, font=\footnotesize\sffamily] at (0,-7.9) {Normalize metrics};
\node[align=left, font=\footnotesize\sffamily] at (0,-8.3) {Generate final reports};

\end{tikzpicture}
\caption{Sequence-style control-flow diagram of the proposed reporting framework.
The main process spawns external monitoring and carbon-emission services and executes
models on the evaluation dataset. During execution, energy consumption $E(m)$ is
measured periodically, while carbon-emission values are retrieved asynchronously.
After execution completes, performance, energy, and carbon metrics are normalized
and aggregated to generate final reports. The framework operates purely as a
reporting layer and does not influence model execution, training, or optimization.}
\label{fig:care-ai-sequence}
\end{figure*}

\section{Standardized Carbon-Aware Evaluation in ML}
Several tools exist to measure or estimate the energy consumption and carbon
emissions of ML workloads, including CodeCarbon, CarbonTracker,
Experiment Impact Tracker, Green Algorithms, and hardware-level power
monitoring utilities \cite{Lacoste2019Quantifying,Anthony2020CarbonTracker,Henderson2020Energy,Lannelongue2021Green}. These tools provide mechanisms
for tracking energy usage and estimating associated carbon emissions during
model training or execution. However, they primarily report raw measurements and
do not prescribe how such information should be normalized, compared, or
interpreted alongside (conventional) task performance metrics. As a result, reported energy and carbon values are often inconsistent, incomplete, or difficult to compare
across studies, limiting their usefulness for deployment-aware model
evaluation.

What is missing is a standardized reporting measurement-agnostic layer that integrates accuracy, energy, and carbon into a coherent and interpretable evaluation view. Carbon-aware evaluation can therefore be viewed as a form of
\emph{deployment-level explainability}, complementing model-centric explainable
artificial intelligence (XAI) methods that aim to clarify how and why model
predictions are produced \cite{wall2025winsorcamhumantunablevisualexplanations}. 
Once again, our goal is not to prescribe optimization strategies, but to enable transparent, comparable, and interpretable reporting of accuracy, energy, and carbon trade-offs to support informed, sustainability-aware model selection \cite{santosh2025carbonneutralhumanairethinking}.




\section{Design and Implementation}



The design philosophy underlying AI-CARE is guided by a set of principles aimed
at enabling standardized, reproducible, and deployment-relevant carbon-aware
evaluation, while minimizing disruption to existing experimental workflows.
These principles are summarized as follows:

\noindent \textbf{Reporting-Centric.} It is designed as a reporting and analysis framework rather than a training-time optimization or control mechanism. The framework does not alter model architectures, learning algorithms, execution schedules, or system
configuration. Instead, it focuses on reporting performance, energy, and carbon
metrics in a unified and comparable form based on observed executions.


\noindent \textbf{Low Overhead.} It is designed so that its runtime overhead is negligible relative to model execution. Energy monitoring and carbon-emission retrieval are performed asynchronously, ensuring that measurement does not interfere with execution
timing, optimization dynamics, or system behavior.

\noindent \textbf{Interpretable.} Raw energy measurements and carbon values are often difficult to interpret in isolation. AI-CARE emphasizes standardized normalization, aggregation, and visualization mechanisms, including Carbon--Accuracy Tradeoff Curves and
composite scores, to support meaningful comparison and decision-making
across models and datasets.

AI-CARE is implemented as a modular reporting framework in which the main
evaluation process coordinates model execution while delegating energy
measurement and carbon-emission retrieval to external services. As illustrated
in Fig.~\ref{fig:care-ai-sequence}, energy consumption is sampled periodically
during execution, while grid carbon-emission values are retrieved
asynchronously at fixed intervals. After execution completes, performance,
energy, and carbon metrics are normalized and aggregated to generate reports, as formalized in Algorithm~\ref{alg:care-ai}.

Unlike predictive approaches that estimate future energy or carbon footprint,
the proposed tool reports empirically observed execution behavior. This design choice
prioritizes measurement transparency, reproducibility, and comparability across
studies, making it suitable as a standardized evaluation layer for
carbon-aware model assessment.

\begin{algorithm}[!t]
\caption{AI-CARE: Carbon-Aware Reporting Evaluation}
\label{alg:care-ai}
\begin{algorithmic}[1]
\REQUIRE Model set $\mathcal{M}$, dataset $\mathcal{D}$, device $d$, weight $\alpha$
\ENSURE Reported carbon--performance tradeoff curves

\FOR{each model $m \in \mathcal{M}$}
    \STATE \textbf{spawn} monitoring service $\mathcal{E}$ on device $d$
    \STATE \textbf{spawn} carbon service $\mathcal{C}$ 
    \STATE $\mathcal{E}.\text{start}$, $\mathcal{C}.\text{start}$ 

    \STATE Execute model $m$ on $\mathcal{D}$ 
    \STATE Compute task performance $P(m)$

    \STATE $\mathcal{E}.\text{stop}()$, $\mathcal{C}.\text{stop}()$
    \STATE Retrieve  $E_{\mathrm{training}}(m)$, $E_{\mathrm{inference}}(m)$ from $\mathcal{E}$

    \STATE Compute carbon emissions:
    \STATE \quad $C_{\mathrm{training}}(m) \leftarrow \sum_k c \cdot E_{\mathrm{training},k}(m)$
    \STATE \quad $C_{\mathrm{inference}}(m) \leftarrow \sum_k c \cdot E_{\mathrm{inference},k}(m)$
    \STATE Compute total carbon:
    \STATE \quad $C_{\mathrm{total}}(m) \leftarrow C_{\mathrm{training}}(m)+C_{\mathrm{inference}}(m)$
\ENDFOR

\STATE Normalize $\{P(m)\}_{m\in\mathcal{M}}$ and $\{C_{\mathrm{total}}(m)\}_{m\in\mathcal{M}}$ 
\STATE Compute \textit{scalar
carbon-aware score} for all $m$:
\STATE \quad $\mathrm{SCAS}(m;\alpha) \leftarrow \alpha \hat{P}(m) + (1-\alpha)\bigl(1-\hat{C}(m)\bigr)$
\STATE Generate \textit{carbon--performance tradeoff curves ($(C_{\mathrm{total}}(m),P(m))$)}
\RETURN Reported metrics and visualizations
\end{algorithmic}
\end{algorithm}
\subsection{Problem Statement}
\label{subsec:formal_quantities}

AI-CARE reports a standardized set of empirical quantities describing predictive
performance, energy consumption, and derived carbon emissions. These quantities
are reported for documentation, transparency, and reproducibility purposes;
it does not optimize or control any of them.

Let $\mathcal{M}$ denotes a set of models evaluated under fixed hardware
and software conditions. 
For a model $m \in \mathcal{M}$, let $P(m) \in [0,1]$ denote a normalized task-dependent performance metric, such as classification accuracy, F1-score, precision, recall, or mean average precision. 


We model computation energy using the number of floating-point operations
(FLOPs), as FLOPs provide a hardware-independent proxy for the amount of
arithmetic work performed by the model $m$. Similarly, memory access counts are used
to capture data movement costs, which are known to be a major contributor to
energy consumption in contemporary CPU and GPU architectures, often exceeding
the cost of arithmetic operations \cite{Horowitz2014Energy}. Static power dissipation accounts for baseline energy draw that is independent of active computation and scales with execution time.

This abstraction mirrors the structure used by practical energy accounting
tools (e.g., CodeCarbon-style monitors \cite{lacoste2019codecarbon}), while remaining sufficiently general to accommodate direct measurement, analytical estimation, or hybrid monitoring approaches.

Formally, for a model $m$ executed for a duration $t$ on device
$d \in \{\mathrm{CPU}, \mathrm{GPU}\}$, total energy consumption is modeled as

\begin{equation}
E(m)
= \alpha_d \cdot \mathrm{FLOPs}(m)
+ \beta_d \cdot \mathrm{Mem}(m)
+ P_d^{\mathrm{static}} \cdot t ,
\label{eq:energy_model}
\end{equation}
where:
\begin{itemize}
    \item $\mathrm{FLOPs}(m)$ denotes the number of floating-point operations,
    \item $\mathrm{Mem}(m)$ denotes the number of memory accesses,
    \item $\alpha_d$ and $\beta_d$ are device-dependent energy coefficients,
    \item $P_d^{\mathrm{static}}$ denotes static power draw.
\end{itemize}

In practice, AI-CARE supports both analytical estimation and empirical measurement. 
When hardware-level monitoring tools are available, energy consumption is 
directly measured during execution. The formulation in 
(\ref{eq:energy_model}) provides a device-aware abstraction that remains 
compatible with analytical, measurement-based, or hybrid energy accounting 
backends.

Energy is accumulated separately for training and inference:
\begin{equation}
\begin{aligned}
&E_{\mathrm{training}}(m) = E_{\mathrm{training}}^{\mathrm{computation}}(m)
+ E_{\mathrm{training}}^{\mathrm{static}}(m), \\
& E_{\mathrm{inference}}(m) = E_{\mathrm{inference}}^{\mathrm{computation}}(m)
+ E_{\mathrm{inference}}^{\mathrm{static}}(m).
\end{aligned}
\end{equation}
Energy values are converted to kilowatt-hours (kWh) via:
\begin{equation}
E_{\mathrm{kWh}}(m) = \frac{E(m)}{3.6 \times 10^6}.
\end{equation}
Given a fixed grid carbon intensity
$c \in \mathbb{R}_{>0}$ (gCO$_2$/kWh), we report carbon emissions following
standard carbon accounting practice \cite{Patterson2021Carbon}:
\begin{equation}
\begin{aligned}
&C_{\mathrm{training}}(m) = c \cdot E_{\mathrm{training}}(m), \\
&C_{\mathrm{inference}}(m) = c \cdot E_{\mathrm{inference}}(m).
\end{aligned}
\end{equation}
For reporting, AI-CARE summarizes empirical tradeoffs using the
\emph{carbon--accuracy tradeoff set}, defined as
\begin{equation}
\mathcal{T} = \{ (C_{\mathrm{total}}(m), P(m)) \mid m \in \mathcal{M} \}.
\end{equation}

Models on the Pareto frontier of $\mathcal{T}$ represent non-dominated tradeoffs
between predictive performance and inference-time carbon cost. 
When a single scalar comparison is required, it reports a \textit{scalar
carbon-aware score (SCAS)}:
\begin{equation}
\mathrm{SCAS}(m;\alpha)
=
\alpha \cdot \hat{P}(m)
+
(1-\alpha)\cdot \bigl(1 - \hat{C}(m)\bigr),
\end{equation}
where $\hat{P}(m)$ and $\hat{C}(m)$ are min--max normalized performance and
carbon values computed over the evaluated model set, and
$\alpha \in [0,1]$ controls the relative emphasis. Unless otherwise specified, we set $\alpha = 0.5$ to assign equal weight 
to predictive performance and carbon impact.

\section{Experiments and Results}
\label{sec:practice}


To evaluate the behavior and performance of the proposed tool, experiments were conducted on five vision benchmarks of increasing complexity:
MNIST \cite{LeCun1998MNIST},
Fashion-MNIST \cite{Xiao2017FashionMNIST},
CIFAR-10 and CIFAR-100 \cite{Krizhevsky2009CIFAR},
and ImageNet-100 derived from ImageNet \cite{Deng2009ImageNet}. All datasets are loaded through a unified data interface. Dataset-specific properties, including input resolution, number of channels, and number of classes, are inferred automatically by the evaluation pipeline at runtime. This design eliminates hard-coded assumptions and ensures consistent evaluation across heterogeneous benchmarks.

We evaluate representative architectures spanning multiple model families, including multilayer perceptrons (MLP), convolutional neural networks (CNNs), compact transformer-based classifiers, and MLP-Mixer \cite{Tolstikhin2021MLPMixer} models that replace attention and convolution with pure multilayer perceptrons for vision tasks. Model hyperparameters are held fixed across datasets in order to isolate architectural and dataset effects rather than gains due to dataset-specific tuning.

All models are trained using identical optimization settings: the Adam optimizer with a learning rate of $10^{-3}$, batch size of 64, and 10 training epochs. A small, fixed number of epochs is intentionally used to emphasize relative accuracy--energy tradeoffs rather than absolute peak accuracy. Inference-time carbon emissions are computed using a fixed grid carbon intensity of 400~gCO$_2$/kWh.
 This value is chosen as a representative average electricity emission factor and is consistent with
recent global estimates of grid carbon intensity reported by international
energy and climate agencies. Recent analyses indicate that global average
electricity emission factors lie in the range of approximately
350--450~gCO$_2$/kWh, reflecting a mixed generation portfolio dominated by
fossil fuels with increasing penetration of renewable energy
\cite{IEA2023Electricity,Ember2023GlobalElectricity}. Using a fixed average carbon-emission factor enables standardized and comparable carbon reporting across models and datasets, independent of geographic location or real-time grid variability. This practice is widely adopted in prior work on carbon-aware ML evaluation and energy
reporting tools, where average grid intensities are preferred for reproducible
benchmarking and cross-study comparison \cite{Lacoste2019Quantifying,Patterson2021Carbon}.
\begin{figure*}[!t]
  \centering
  \includegraphics[width=1.08\textwidth]{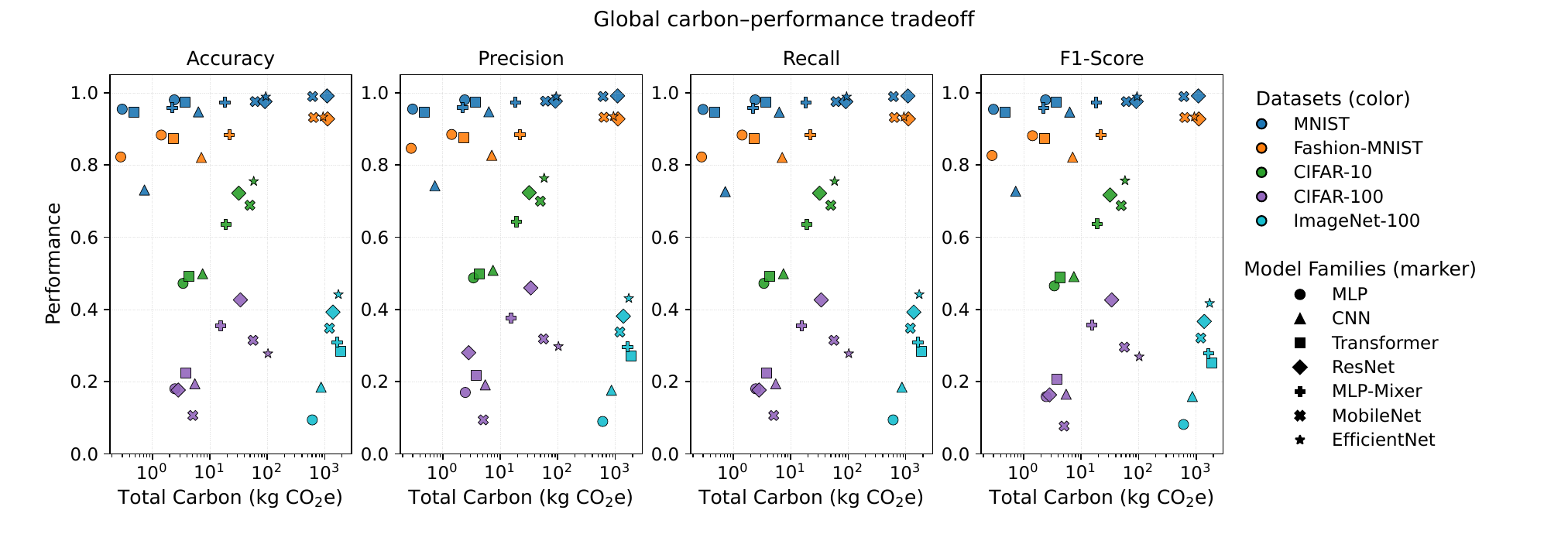}

  \caption{
  Global carbon--performance tradeoff across all evaluated model--dataset pairs.
  Each subplot reports task performance (Accuracy, Precision, Recall, and F1-Score)
  versus total carbon emissions (training + inference) shown on a logarithmic scale.
  Colors indicate datasets, while marker shapes denote model families.
  }
  \label{fig:catc_all_metrics}
\end{figure*}

\begin{figure*}[!t]
  \centering
  \includegraphics[width=0.91\textwidth]{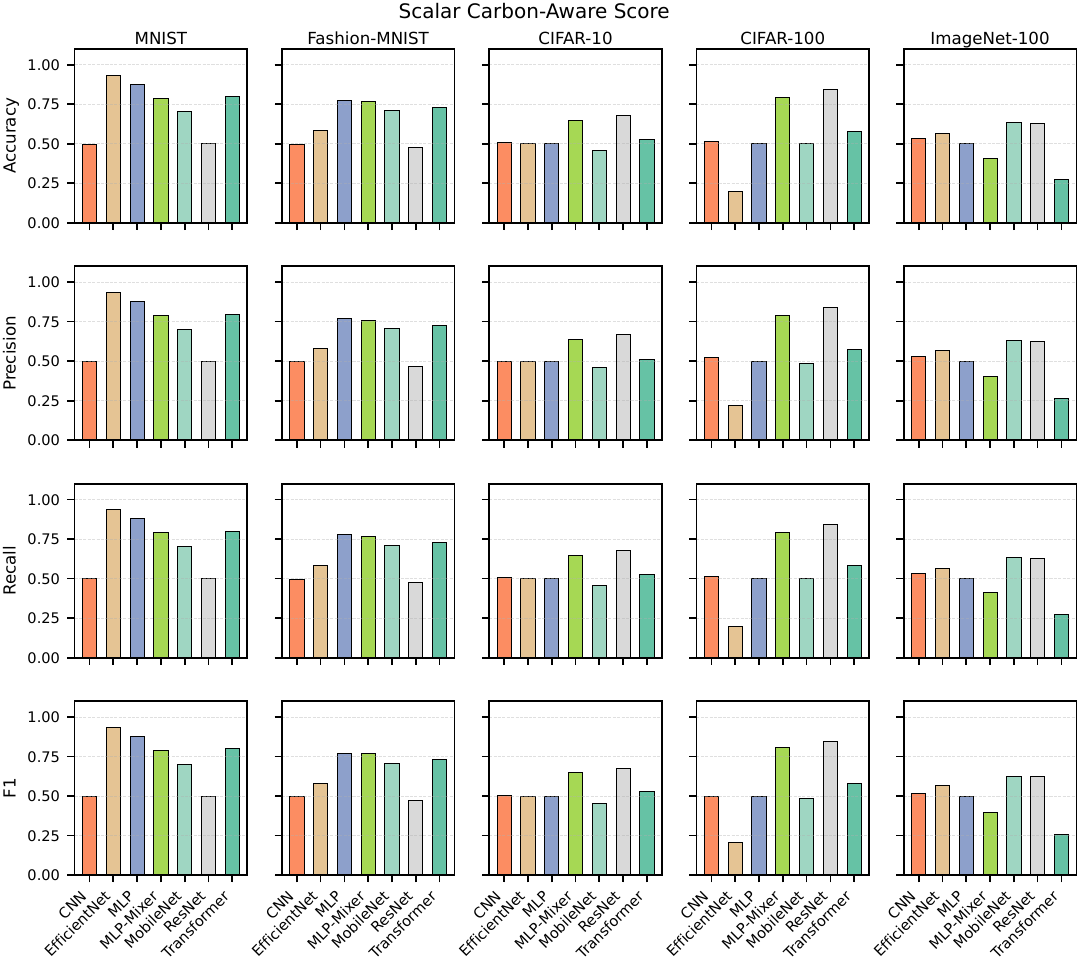}
  \caption{
  Metric-wise scalar carbon-aware scores across datasets and model families.
  Rows correspond to evaluation metrics (accuracy, precision, recall, and F1-score),
  while columns correspond to datasets (MNIST, Fashion-MNIST, CIFAR-10, CIFAR-100,
  and ImageNet-100).
  Each score integrates normalized task performance with total carbon emissions
  (training + inference). The grouped layout enables direct cross-metric and
  cross-dataset comparison of carbon-aware model rankings.
  }
  \label{fig:care_ai_metric_multipanel}
\end{figure*}

Figure~\ref{fig:catc_all_metrics} provides a global view of empirical
carbon--performance tradeoffs across all evaluated model--dataset pairs.
Each point corresponds to a trained model evaluated under consistent
experimental settings, plotted against total carbon emissions
(training + inference) on a logarithmic scale. The four panels report
Accuracy, Precision, Recall, and F1, enabling a multi-metric assessment
of efficiency beyond single-score evaluation. Colors indicate datasets,
while marker shapes denote model families. For example, on CIFAR-100, deeper architectures achieve only marginal 
performance improvements relative to compact CNN models, while incurring 
multiple-fold increases in total carbon emissions. Such disparities become 
explicit in the carbon--performance tradeoff plots.

Across all metrics, a consistent structural pattern emerges. As dataset
complexity increases, improvements in predictive performance are
accompanied by disproportionately higher carbon cost. In particular,
the upper-right regions of the plots are populated by deeper or more
computationally intensive architectures that yield incremental
performance gains at substantially greater carbon expense. Conversely,
several models cluster near the lower-left frontier, indicating
favorable tradeoffs in which competitive performance is achieved at
markedly lower total emissions.

Importantly, the multi-metric perspective reveals that carbon
inefficiency is not confined to accuracy alone. Models with comparable
Accuracy may exhibit non-trivial differences in Precision, Recall, or
F1, while simultaneously differing by orders of magnitude in total
carbon emissions. Such discrepancies remain obscured under
performance-only benchmarking.

The carbon--performance curves therefore make efficiency disparities
explicit, enabling identification of Pareto-efficient configurations and
supporting deployment-aware model selection. By jointly considering
predictive quality and environmental cost, this analysis reinforces the
need for carbon-aware evaluation frameworks in modern AI systems.

Figure~\ref{fig:care_ai_metric_multipanel} presents scalar carbon-aware scores
(CARE-AI) across four evaluation metrics (accuracy, precision, recall,
and F1-score) and five benchmark datasets (MNIST, Fashion-MNIST,
CIFAR-10, CIFAR-100, and ImageNet-100).
Rows correspond to evaluation metrics, while columns correspond to datasets,
enabling direct cross-metric and cross-dataset comparison within a unified layout.
Each score jointly normalizes predictive performance and total carbon emissions
(training + inference), providing a sustainability-aware ranking of model families.

Across metrics, the results reveal that carbon-aware rankings are influenced
not only by energy consumption but also by the choice of performance measure.
Accuracy and F1-score generally produce consistent ordering patterns,
whereas precision- and recall-based evaluations alter rankings in several cases.
This variation reflects differences in prediction behavior and class-wise
error distributions, particularly for more complex datasets.

Across datasets, a clear complexity-driven trend emerges.
On simpler benchmarks such as MNIST and Fashion-MNIST,
lightweight architectures consistently achieve high scalar carbon-aware scores,
indicating that competitive performance can be obtained at minimal carbon cost.
In these settings, larger architectures yield limited relative gains in
predictive performance and are penalized by their higher energy expenditure.

As task complexity increases from CIFAR-10 to CIFAR-100 and ImageNet-100,
efficiency trade-offs become substantially more pronounced.
Incremental improvements in predictive performance often require
disproportionately higher carbon expenditure,
resulting in lower scalar carbon-aware scores for energy-intensive models.
Conversely, architectures with slightly lower raw performance but significantly
reduced carbon cost frequently achieve more favorable sustainability rankings.

Overall, the grouped metric–dataset view highlights that AI-CARE provides
a fundamentally different evaluation perspective than conventional
performance-only benchmarking.
By integrating normalized task performance with carbon impact,
the scalar carbon-aware score exposes deployment-relevant efficiency trade-offs
that remain invisible under traditional evaluation frameworks.


Across all benchmarks and evaluation metrics, the combined analysis
demonstrates that carbon-aware evaluation fundamentally reshapes model
assessment. The carbon--performance tradeoff plots reveal consistent,
nonlinear efficiency disparities: incremental gains in predictive
quality frequently require disproportionately higher carbon expenditure,
particularly as dataset complexity increases. These disparities become
more pronounced in higher-complexity benchmarks such as CIFAR-100 and
ImageNet-100, where deeper architectures cluster in high-emission
regions while offering only marginal performance improvements.

The grouped CARE-AI scalar scores complement this geometric analysis by
providing a unified, cross-metric ranking framework. By jointly
normalizing predictive performance and total carbon emissions, the
scalar formulation exposes metric-dependent variations in model
ordering and highlights architectures that achieve favorable
performance--emissions balance across datasets. Importantly, models
that appear comparable under accuracy-only evaluation can diverge
substantially once carbon cost and alternative metrics are considered.

Together, the tradeoff curves and scalar scores provide structured,
deployment-relevant evidence that energy and carbon impact materially
influence empirical conclusions about model quality. The results
underscore the necessity of standardized carbon-aware reporting and
demonstrate that sustainability should be treated as a first-class
evaluation dimension alongside predictive performance in modern AI
systems.







\section{Conclusion}

This paper introduced AI-CARE, a standardized framework for carbon-aware
evaluation of ML models. By jointly reporting predictive
performance, energy consumption, and carbon emissions, AI-CARE reveals
efficiency tradeoffs that remain obscured under conventional
accuracy-centric benchmarking. Through carbon--performance tradeoff
curves and scalar carbon-aware scores computed across multiple evaluation
metrics, the framework provides both geometric and ranking-based
perspectives on sustainability-aware model assessment.

Empirical results across diverse vision benchmarks demonstrate that
performance improvements frequently incur disproportionately higher
carbon cost, particularly as dataset complexity increases. Moreover,
models that appear comparable under single-metric evaluation can diverge
substantially when carbon impact and alternative performance measures
are considered. The multi-metric analysis further shows that carbon-aware
rankings are sensitive to evaluation criteria, reinforcing the importance
of comprehensive reporting.

Together, these findings highlight the limitations of single-objective
model evaluation and underscore the need for transparent, standardized
multi-objective reporting practices. AI-CARE provides a practical tool
to support deployment-aware model selection in resource- and
climate-conscious environments. We hope this work contributes to the
broader integration of energy and carbon metrics as first-class
evaluation criteria in modern AI research and practice.




\section*{Artifact Availability}
The AI-CARE tool is publicly available as open-source software and reproduces
all reporting procedures described in this document. The tool and documentation are available at \url{https://github.com/USD-AI-ResearchLab/ai-care}.

\section*{Acknowledgment}
This work was supported by the National Science Foundation under Grant No. \href{https://www.nsf.gov/awardsearch/showAward?AWD_ID=2346643}{\#2346643}, the U.S. Department of Defense under Award No. \href{https://dtic.dimensions.ai/details/grant/grant.14525543}{\#FA9550-23-1-0495}, and the U.S. Department of Education under Grant No. P116Z240151.
Any opinions, findings, conclusions or recommendations expressed in this material are those of the author(s) and do not necessarily reflect the views of the National Science Foundation, the U.S. Department of Defense, or the U.S. Department of Education.

\balance
\bibliographystyle{IEEEtran}
\bibliography{references}

\end{document}